\documentclass{article}
\usepackage{spconf,amsmath,graphicx}
\usepackage[font=small,labelfont=bf]{caption}
\usepackage{subcaption}
\usepackage{booktabs}
\usepackage{derivative}
\usepackage[dvipsnames]{xcolor}
\usepackage{color, colortbl}
\usepackage{pbox}
\definecolor{Gray}{gray}{0.9}
\usepackage{amssymb}

\def\ie{\emph{i.e.}}
\def\eg{\emph{e.g.}}
\def\etal{\emph{et al.~}}

\title{Hearing Loss Detection from Facial Expressions in \\ One-on-one Conversations}

\name{
\begin{tabular}{@{}c@{}}
Yufeng Yin$^{1*}$\thanks{*Work done during an internship at Meta Reality Labs Research.}, 
Ishwarya Ananthabhotla$^2$, 
Vamsi Krishna Ithapu$^2$ \\ 
Stavros Petridis$^3$, 
Yu-Hsiang Wu$^{2,4}$, 
Christi Miller$^2$
\end{tabular}}

\address{$^1$ University of Southern California, USA, $^2$ Meta Reality Labs Research, USA,\\$^3$ Meta, UK, $^4$ University of Iowa, USA}

\begin{document}
\ninept

\maketitle

\begin{abstract}
\noindent Individuals with impaired hearing experience difficulty in conversations, especially in noisy environments. This difficulty often manifests as a change in behavior and may be captured via facial expressions, such as the expression of discomfort or fatigue. 
In this work, we build on this idea and introduce the problem of detecting hearing loss from an individual's facial expressions during a conversation. Building machine learning models that can represent hearing-related facial expression changes is a challenge. In addition, models need to disentangle spurious age-related correlations from hearing-driven expressions. 
To this end, we propose a self-supervised pre-training strategy tailored for the modeling of expression variations. We also use adversarial representation learning to mitigate the age bias. 
We evaluate our approach on a large-scale egocentric dataset with real-world conversational scenarios involving subjects with hearing loss and show that our method for hearing loss detection achieves superior performance over baselines.
\end{abstract}


\section{Introduction}
Hearing loss refers to a partial or total inability to hear sounds in one or both ears \cite{nadol1993hearing}. Common causes of hearing loss include: age, noise exposure, genetics, etc \cite{daniel2007noise}. Hearing loss impacts effective communication and interpersonal relationships, especially during social interactions \cite{matkin1999considerations, lohler2019hearing}. However, traditional assessment of hearing loss often fail to capture these aspects of hearing loss, instead focusing on medical aspects of hearing loss or using retrospective surveys. Similarly, previous studies of machine learning-based hearing loss detection collect input data from noise exposure duration \cite{zhao2019machine}, micro RNA \cite{shew2019using}, and screening tests \cite{lenatti2022evaluation} which are not directly accessible in social interaction and thus can not be leveraged for real-time inference. Timely detection of more nuanced, functional aspects of hearing loss allows for appropriate interventions such as communication strategies or hearing aids, enabling individuals to actively engage in conversations.

\begin{figure}[t]
\centering
\includegraphics[trim=26 26 0 26, clip, width=0.4\textwidth]{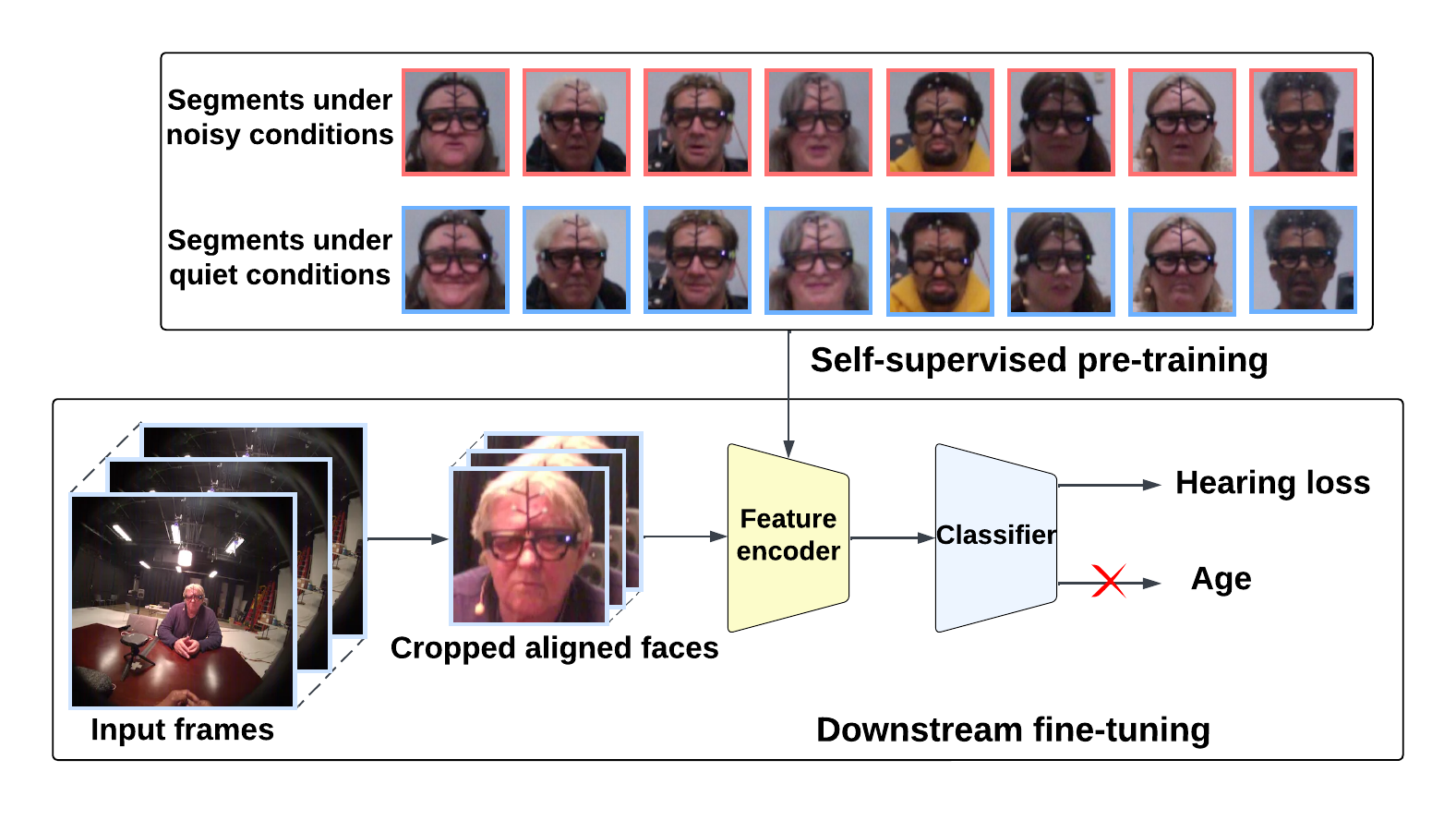}
\vspace{-5pt}
\caption{We study the novel problem of hearing loss detection from facial expressions in one-on-one conversations. Given a video clip recording the subject's facial expressions in one-on-one conversations, we detect if the subject has hearing loss. Our method learns the within-subject variability of facial expressions via self-supervised pre-training while reducing the age bias in downstream fine-tuning.}
\vspace{-10pt}
\label{fig:teaser}
\end{figure}

In this paper, we introduce the novel problem of using facial expressions for real-time hearing loss detection in one-on-one conversations (see Fig. \ref{fig:teaser}). 
There are two main challenges: (i) We need to model the complex relationships between an individual's hearing ability and facial behaviors. (ii) Hearing loss has a strong correlation with age. Thus, machine learning models may use age as a shortcut for hearing loss detection which hurts the performance, especially for young subjects. Previous studies show that people with hearing loss have more difficulty in listening under noisy conditions than those with normal hearing \cite{heinonen2005genetic, heinonen2011noise}. 
In addition, studying the range of facial feature variability in a given subject is critical for effective expression recognition \cite{yang2019learning, verburg2019micro}. Thus, we design a method that leverages variability in facial expressions across background noise levels and explicitly attempts to counteract the age bias present in the data to achieve hearing loss detection.

The major contributions of our work are: (i) We introduce the novel problem of hearing loss detection from facial expressions in one-on-one conversations. (ii) We use self-supervised pre-training to learn representations that describe the variation of facial expressions within a single subject. (iii) We find that age bias in the data hurts the model's performance for younger subjects, and propose to mitigate it with adversarial representation learning. (iv) We evaluate our approach on a naturalistic, egocentric dataset and show the effectiveness of our proposed model.

\begin{figure*}[t]
\centering
\includegraphics[trim=23 18 21 22, clip, width=0.95\textwidth]{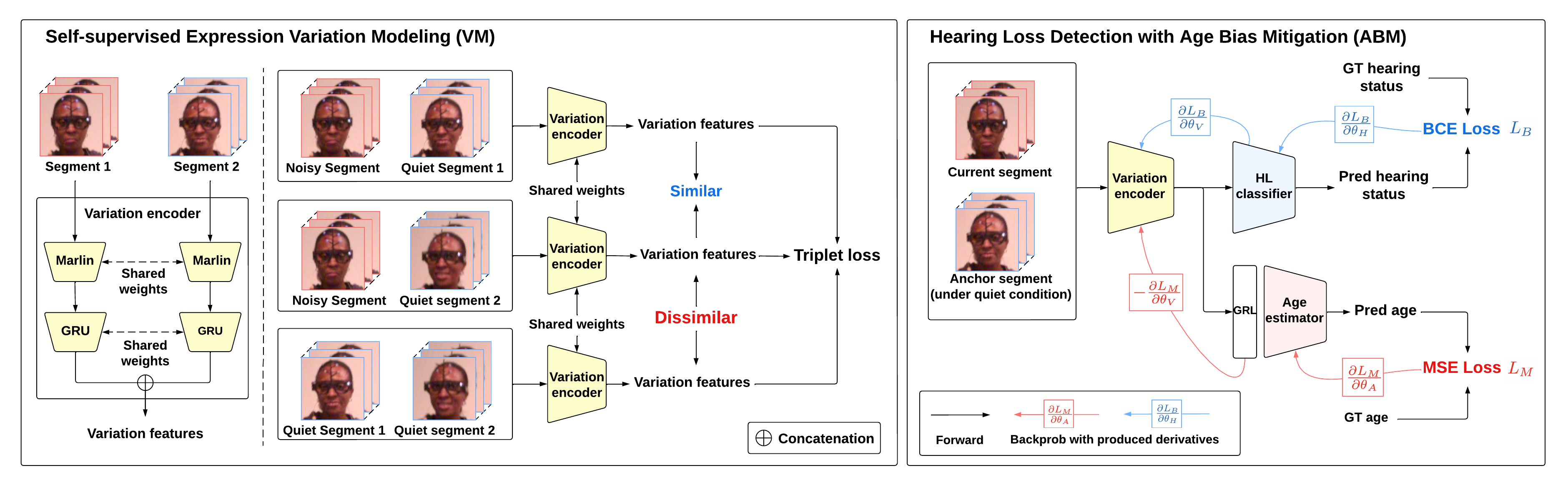}
\vspace{-5pt}
\caption{(Best viewed in color) Architectural overview of the proposed method for hearing loss detection in one-on-one conversations: (i) We pre-train the encoder with expression variation modeling to capture the feature variations across noise levels. (ii) We fine-tune the model for hearing loss detection with age bias mitigation.}
\vspace{-10pt}
\label{fig:overview}
\end{figure*}

\section{Related work}
\noindent \textbf{Hearing loss detection.}
Recently, there has been a growing interest in leveraging machine learning techniques for hearing loss detection \cite{zhao2019machine, shew2019using, lenatti2022evaluation, tomiazzi2019performance}. 
Zhao \etal \cite{zhao2019machine} find that machine learning is a potential tool for the detection of noise-induced hearing loss when age and exposure duration are used as inputs. 
Shew \etal \cite{shew2019using} develop machine learning models using micro RNA (miRNA) to predict the presence of hearing loss. 
Lenatti \etal \cite{lenatti2022evaluation} investigate the contributions of the measured features in machine learning models toward hearing loss detection in speech-in-noise screening tests. The results demonstrate that age, number of correct responses, and average reaction time have high relevance with hearing loss. 
Although promising performance in hearing loss detection is found, the information required in the previous studies is not directly accessible in social interactions, making real-time inference impossible. Motivated by these findings, in this paper, we explore using non-verbal behaviors in social interactions, \eg, facial expressions, for real-time hearing loss detection.

\noindent \textbf{Visual cues in communication difficulties.}
Support for this line of work comes from prior evidence for visual cues in difficult communication situations, \eg, with background noise or hearing loss \cite{hadley2019speech, hadley2021synchrony, venkitakrishnan2023facial}. 
Hadley \etal \cite{hadley2019speech} find that when the background noise level increases, people tend to speak louder (mouth wide open) and move closer together. In addition, they observe increased eye gaze on the speaker's mouth for lip reading. 
Later, Hadley \etal \cite{hadley2021synchrony} report increased head movement coherence in high background noise levels.
Finally, Venkitakrishnan \etal \cite{venkitakrishnan2023facial} observe that facial expressions of confusion and frustration (particularly, brow lowerer \cite{ekman1977facial}) increase as communication difficulty increases. These results support the potential use of facial expressions as a measure of affect and listening difficulty.

\section{Method}
\subsection{Problem Formulation}
Given a video set $S$, each video segment $s \in S$ records the subject's facial expressions in one-on-one conversations. We aim to detect the subject's hearing status $h = \text{F}(s)$, where $h$ is a binary label. $h = 1$ if the subject has hearing loss otherwise $h = 0$.

\subsection{Model}
We present the model overview in Fig. \ref{fig:overview}. The variation encoder is pre-trained to capture the expression variations within a subject across noise levels with self-supervised learning. Then, the whole model is fine-tuned for hearing loss detection with age bias mitigation by adversarial representation learning. During inference, only the current and anchor segments (the first segment under the quiet condition in the session) are used as input, while age or noise level information is not required.

\subsubsection{Facial Feature Encoding}
We use Marlin \cite{cai2022marlin} to extract the facial features for the input videos. Marlin is a facial representation model for videos with a transformer architecture \cite{vaswani2017attention}. It is pre-trained with mask modeling \cite{he2022masked} on YouTube Faces \cite{wolf2011face} which collects web-crawled facial videos from YouTube. Marlin learns strong facial representations from videos which can be applied to a variety of downstream tasks for face, \eg, facial attribute recognition \cite{zhu2022celebv} and facial expression recognition \cite{li2020deep}. In particular, we use Marlin Small for efficient training and inference. Then, the temporal features are passed to a one-layer Gated Recurrent Unit (GRU) with average pooling. 
Formally, given a video segment $s$, we extract the feature $f$ with the facial encoder $\text{E}$, which consists of a Marlin encoder and a GRU: $f = \text{E}(s)$.

\subsubsection{Self-supervised expression variation modeling}
\noindent \textbf{Variation encoding.} 
Given two segments from the same subject but under different background noise levels, \ie, $s_1$ and $s_2$, to obtain the facial feature variability between them, we extract the facial features with the same facial feature encoder $\text{E}$ and then concatenate them together: $v = \text{Concat}(\text{E}(s_1), \text{E}(s_2))$. The entire process of variation encoding is denoted as $\text{V}$.

\noindent \textbf{Variation modeling.} 
As discussed previously, individuals with hearing loss have more difficulty listening under noisy conditions than those with normal hearing. This difference is expected to manifest as changes in listeners' facial expressions, but the variability is also expected to be subject-dependent. We leverage this idea in a self-supervised pre-training stage for the variation modeling. In particular, to construct each training instance, we first randomly sample three segments $s_n$, $s_{q1}$, $s_{q2}$ from the same person. $s_n$ is a segment under the noisy condition while $s_{q1}$ and $s_{q2}$ are under the quiet condition. We form three segment pairs from these three segments for variation encoding, in which $v_a = \text{V}(s_n, s_{q1})$ and $v_p = \text{V}(s_n, s_{q2})$ have the same noise level changes while $v_n = \text{V}(s_{q1}, s_{q2})$ has a different noise level change with the former two pairs. We use a triplet loss \cite{hoffer2015deep} to make the positive pair ($v_a$ and $v_p$) closer and the negative pair ($v_a$ and $v_n$) farther apart in the learned embedding space.
\begin{equation}
    \mathcal{L}_{T} = \max\{1+\text{D}(v_a, v_p)-\text{D}(v_a, v_n), 0\},
\end{equation}
\noindent where $\text{D}$ is the Euclidean distance.

\subsubsection{Hearing loss detection with age bias mitigation}
We fine-tune the model for hearing loss detection while mitigating the age bias in the variation encoder. The model consists of three components: variation encoder, hearing loss (HL) classifier, and age estimator. The variation encoder and HL classifier are optimized with binary cross-entropy (BCE) loss: $\mathcal{L}_{B} = \text{BCE}(h, \hat{h})$, where $h$ and $\hat{h}$ denote the ground-truth and prediction for hearing status.

To reduce the age bias in the variation encoder, we employ an adversarial approach: the age estimator is trained to accurately predict the subject's age within a segment (minimizing the age estimation loss) while the variation encoder is trained to confuse the age estimator (maximizing the age estimation loss). The age estimation loss is calculated as the mean squared error (MSE) between the ground-truth age $a$ and predicted age $\hat{a}$: $\mathcal{L}_{M} = \text{MSE}(a, \hat{a})$. 
To implement this, a Gradient Reversal Layer (GRL) \cite{ganin2016domain} is introduced at the beginning of the age estimator. During the forward pass, GRL has no effect, but during the backward pass, GRL reverses the gradient of the loss function by multiplying it by $-1$. Before GRL comes into play, within the age estimator (see Fig. \ref{fig:overview}, gradients from MSE loss to the age estimator), the network strives to minimize the age estimation loss, aligning itself with the goal of accurate age prediction. After GRL's introduction, within the variation encoder (see Fig. \ref{fig:overview}, gradients from GRL to the variation encoder), the gradients are reversed (multiplied by $-1$). Consequently, the variation encoder is updated to maximize the age estimation loss which means reducing the age information.

Overall, the approach uses one-stage training with multi-task learning: $\mathcal{L} = \mathcal{L}_{B} + \mathcal{L}_{M}$.

\section{Experiments}
\begin{figure}[t]
\centering
\footnotesize
    \begin{subfigure}{0.22\textwidth}
    \includegraphics[width=\textwidth]{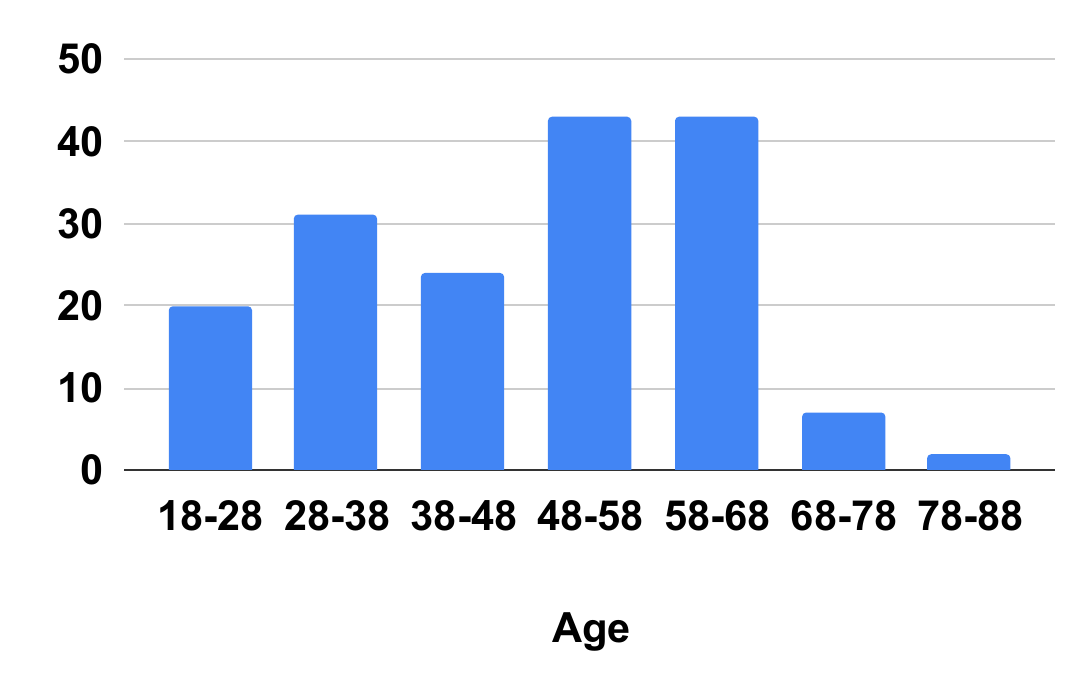}
    \vspace{-10pt}
    \caption{Number of subjects.}
    \vspace{-5pt}
    \label{fig:age_count}
    \end{subfigure}
    \begin{subfigure}{0.22\textwidth}
    \includegraphics[width=\textwidth]{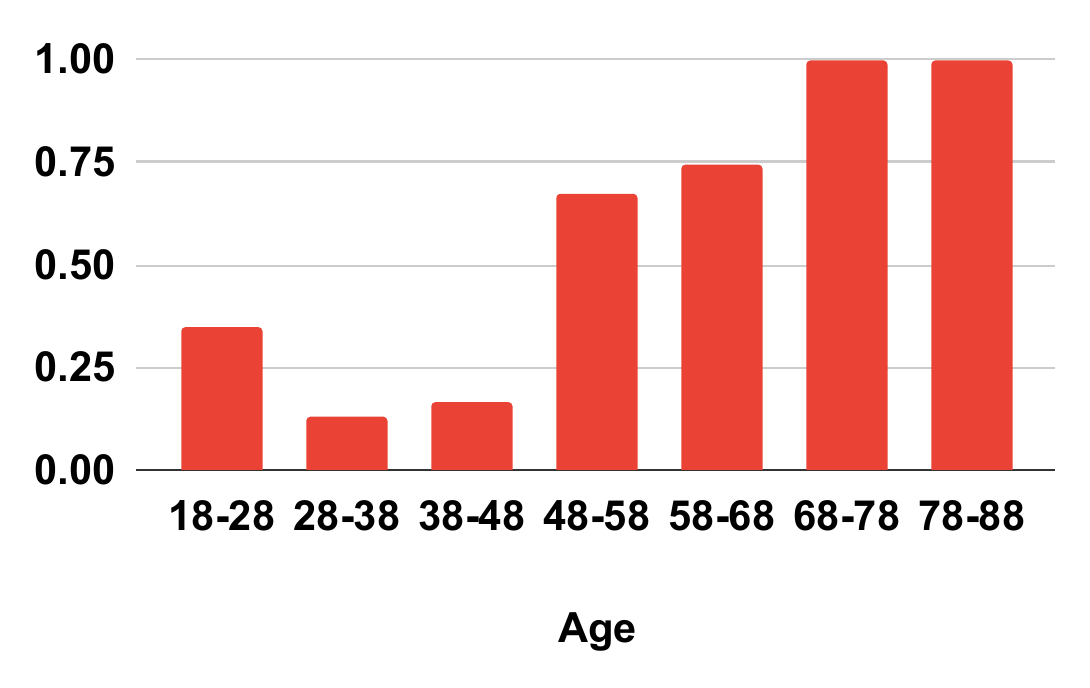}
    \vspace{-10pt}
    \caption{Positive rates.}
    \vspace{-5pt}
    \label{fig:positivate_rate}
    \end{subfigure}
    \caption{Number of subjects and positive rates in different age ranges.}
    \label{fig:data_observation}
\end{figure}

\begin{table}[t]
\centering
\footnotesize
\vspace{-5pt}
\caption{Statistics of the three age groups. They have similar number of subjects while older group has higher positive rate.}
\vspace{-5pt}
\scalebox{1}{
\begin{tabular}{l|p{25pt}|p{25pt}|p{25pt}}
\toprule
\rowcolor{Gray}
Group & Young & Mid. & Old \\
\midrule
Age range & 18-39 & 40-56 & 57-88 \\
\# subjects & 60 & 64 & 54 \\
Positive rate & 0.18 & 0.52 & 0.76 \\
\bottomrule
\end{tabular}}
\vspace{-10pt}
\label{tab:age}
\end{table}

\begin{table}[t]
\centering
\footnotesize
\vspace{-5pt}
\caption{F1 scores (\%, $\uparrow$) for hearing loss detection. We get the highest performance with all the components. * means the corresponding result is significantly better than the previous method with $p<0.1$ by one-tailed $t$-test. P-values are adjusted for multiple comparisons.}
\vspace{-5pt}
\scalebox{0.85}{
\begin{tabular}{l|l|lll}
\toprule
\rowcolor{Gray}
Method & 5-fold CV & Young & Mid. & Old \\
\midrule
Random guessing & 47.9 & 27.5 & 47.6 & 60.2 \\
\midrule
Marlin & 63.4 $\pm$ 1.4 & 32.9 $\pm$ 2.5 & 60.3 $\pm$ 2.5 & 79.3 $\pm$ 1.6 \\
Marlin+Age & 69.9 $\pm$ 0.9 & 16.1 $\pm$ 4.3 & 60.6 $\pm$ 1.9 & 85.7 $\pm$ 1.4 \\
\midrule
Marlin & 63.4 $\pm$ 1.4 & 32.9 $\pm$ 2.5 & 60.3 $\pm$ 2.5 & 79.3 $\pm$ 1.6 \\
Marlin+Anchor & 63.5 $\pm$ 1.0 & 41.1 $\pm$ 1.9$^{*}$ & 59.7 $\pm$ 3.9 & 76.7 $\pm$ 3.2 \\
Marlin+VM & 66.4 $\pm$ 1.2$^{*}$ & 40.6 $\pm$ 2.0 & 59.4 $\pm$ 1.1 & \textbf{80.6 $\pm$ 1.4}$^{*}$ \\
\textbf{Marlin+VM+ABM} & \textbf{66.9 $\pm$ 0.8} & \textbf{44.8 $\pm$ 2.6}$^{*}$ & \textbf{64.3 $\pm$ 2.4}$^{*}$ & 79.4 $\pm$ 2.2 \\
\bottomrule
\end{tabular}}
\vspace{-10pt}
\label{tab:main_exp}
\end{table}

\subsection{Dataset}
To our knowledge, there is no existing face dataset with a diverse subject pool and hearing status labels. To evaluate our approach, we use a subset of spontaneous one-on-one conversations selected from a larger dataset that includes multiparty conversations, \ie, the Reality Labs Research Conversations for Hearing Augmentation Technology (RLR-CHAT) dataset. Participants provide written consent after being informed of study procedures, risks, and data use details. The subset consists of 89 one-on-one sessions and each session is approximately one hour. Every subject wears a pair of data collection glasses\footnote{https://about.meta.com/realitylabs/projectaria/}, which includes an RGB fisheye camera recording the scene with a frame rate of 5 frames per second. During the conversation, eight loudspeakers surround the subjects playing cafeteria noise. The noise level pseudo-randomly changes every 25-35 seconds. There are four noise levels, \ie, no noise (quiet), 55, 65, and 75 dBA, which cover the range of listening conditions experienced in real-world situations \cite{wu2018characteristics}. The noise level distribution is balanced.

Hearing status is established by an audiometric hearing screening test. In particular, the hearing screening is administered in a quiet room using a calibrated audiometer with testing conducted by trained research staff. The subjects are screened at 25 dB HL using interrupted pure tones at 0.5, 1, 2, 3, and 4 kHz in each ear separately. Subjects are considered to be hearing impaired for this study if they fail to hear tones at any one frequency at 25 dB HL. 
Age information for each subject is self-reported. In total, 178 subjects participate in one-on-one sessions, and 85 are labeled as hearing impaired per the failed hearing screening test. The positive rate of hearing loss is 0.48. Fig. \ref{fig:data_observation} shows the data observation of age and positive rate for hearing loss.

We segment the whole session into short video clips based on the noise level changes, resulting in 25-35-second video clips. In total, 18,145 segments with associated hearing status, noise level, and age information are under evaluation. It is worth noting that the hearing status of the subject in the scene video is being predicted, instead of the subject wearing the glasses that records the video. Inferring hearing status from the facial expressions of the glasses-wearer is out of scope for this project.

\subsection{Experiment Setup}
The evaluation metric is the F1 score. Subject-independent 5-fold cross-validation is performed to evaluate the overall performance. Specifically, all sessions are partitioned into five folds. In each iteration, four folds are used for training and one fold is used for testing.
We also evaluate the performance of subjects of different age ranges. In particular, the subjects are evenly divided into three groups, \ie, young, middle-aged (mid.), and old, and then the performance is evaluated separately for each group. Tab. \ref{tab:age} provides the statistics for the three age groups. 
For both evaluations, we train and evaluate with five different random seeds to obtain the performance standard deviations.

\subsection{Implementation Details}
\noindent \textbf{Facial feature extraction.}
We use 2D-FAN \cite{bulat2017far} and FFHQ-alignment for face cropping and alignment. Each resulting image is resized into $224\times224$, and fed to the pre-trained Marlin encoder. The Marlin encoder remains frozen to prevent over-fitting.

\noindent \textbf{Model training.}
We train the models with the AdamW optimizer for 15 epochs with a learning rate of $1e^{-4}$ and a batch size of $64$. To make the scales of the BCE and MSE loss comparable, the age distribution is normalized to 0 mean and 1 standard deviation.

\subsection{Methods}
We use the following model variants in our experiments:

\noindent \textbf{Marlin}: We input the target segment into the Marlin encoder with a GRU layer (facial feature encoding) and then feed the features into the hearing loss classifier.

\noindent \textbf{Marlin+Age}: We concatenate the age information with the encoded facial features of the target segment and input them into the hearing loss classifier.

\noindent \textbf{Marlin+Anchor}: We extract features of the target segment and an anchor segment under quiet conditions with facial feature encoding. Features of the two segments are concatenated and then passed into the hearing loss classifier.

\noindent \textbf{Marlin+VM}: Based on Marlin+Anchor, we first pre-train the variation encoder with variation modeling and then fine-tune the variation encoder and hearing loss classifier for hearing loss detection.

\noindent \textbf{Marlin+VM+ABM}: Based on Marlin+VM, after pre-training with variation modeling, we fine-tune the model for hearing loss detection with the proposed age bias mitigation.

\begin{figure}[t]
\centering
\includegraphics[width=0.32\textwidth]{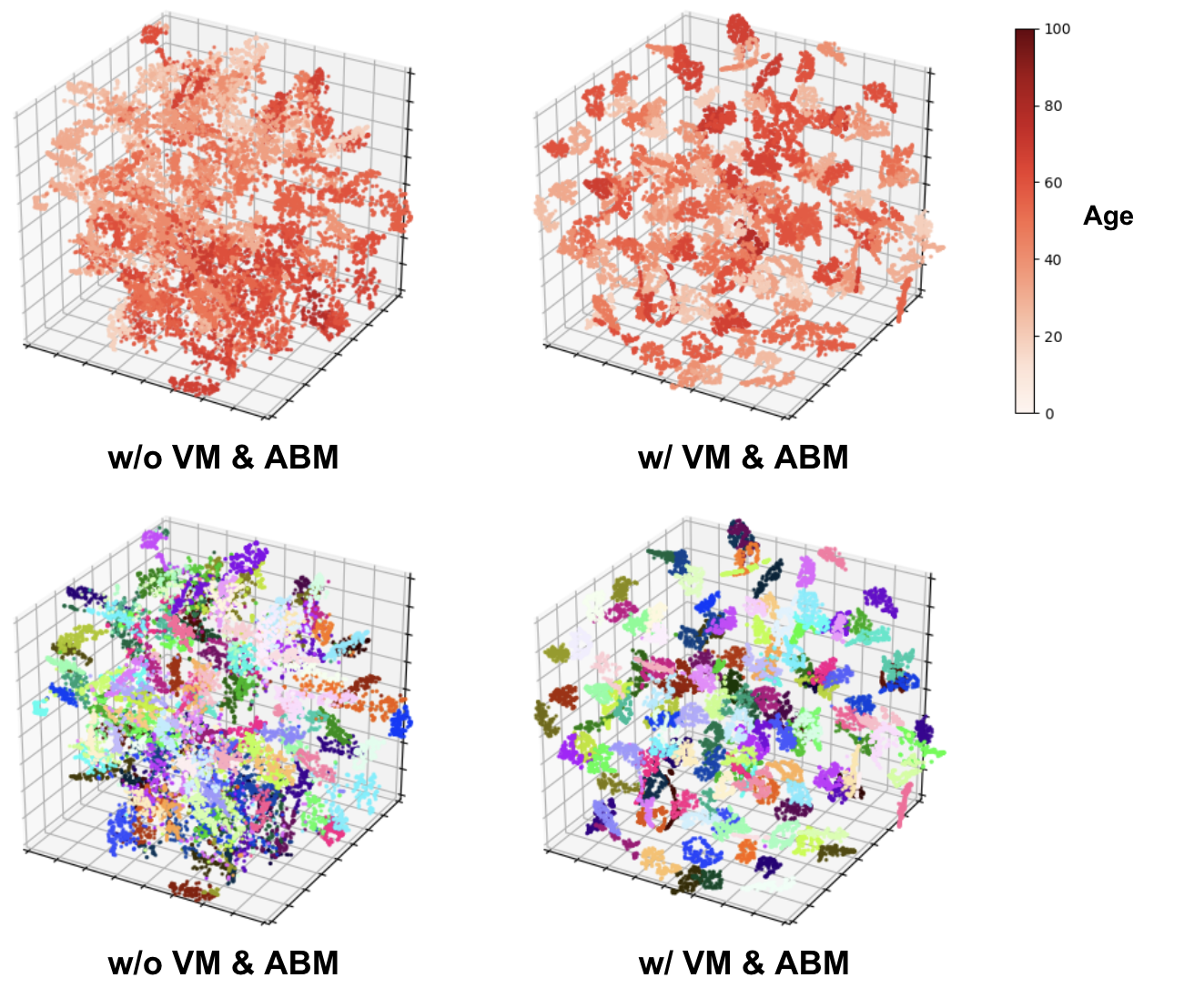}
\vspace{-5pt}
\caption{Visualizations of 3D T-SNE. Each point represents a segment. In the top row, color represents the age while in the bottom row, color represents the identity. With variation modeling and age bias mitigation, segments of different ages are mixed up together while points from the same person are grouped together.}
\vspace{-10pt}
\label{fig:visualization}
\end{figure}

\subsection{Experiment Results}
\noindent \textbf{Age influence.} In Tab. \ref{tab:main_exp}, comparing Marlin and Marlin+Age, we find that adding age improves the overall performance. However, the performance for the young group is decreased and is even worse than random guessing. Thus, age information is a bias in hearing loss detection. The finding motivates the approach of unlearning age.

\noindent \textbf{Model ablation.} Tab. \ref{tab:main_exp} shows the model ablation and the significance test. Adding the anchor segments, variation modeling, and age bias mitigation progressively improve the model performance. Especially, the model achieves the highest performance for young, and middle-aged groups with all the components included. The results indicate that the proposed model learns discriminative information for hearing loss while reducing the age bias.

\noindent \textbf{Visualization.} Fig. \ref{fig:visualization} displays the 3D T-SNE visualizations \cite{van2008visualizing} of features from the variation encoder, with each point representing a segment. In the top row, color indicates age, with darker red denoting older subjects. In the bottom row, color signifies subject identity. 
In the top-left plot (w/o VM \& ABM), there is a trend of color darkening from top to bottom, suggesting age-related clustering. Conversely, in the top-right plot (w/ VM \& ABM), points of different ages are mixed up, indicating the successful removal of age information. 
Furthermore, in the bottom-right plot, segments from the same person are grouped together while the relatedness between the subjects is not dominated by age. The plots show that the model accounts for individual differences in hearing loss detection and reduces the correlation of the representation with the subjects' age.

\begin{figure}[t]
\centering
\includegraphics[width=0.25\textwidth]{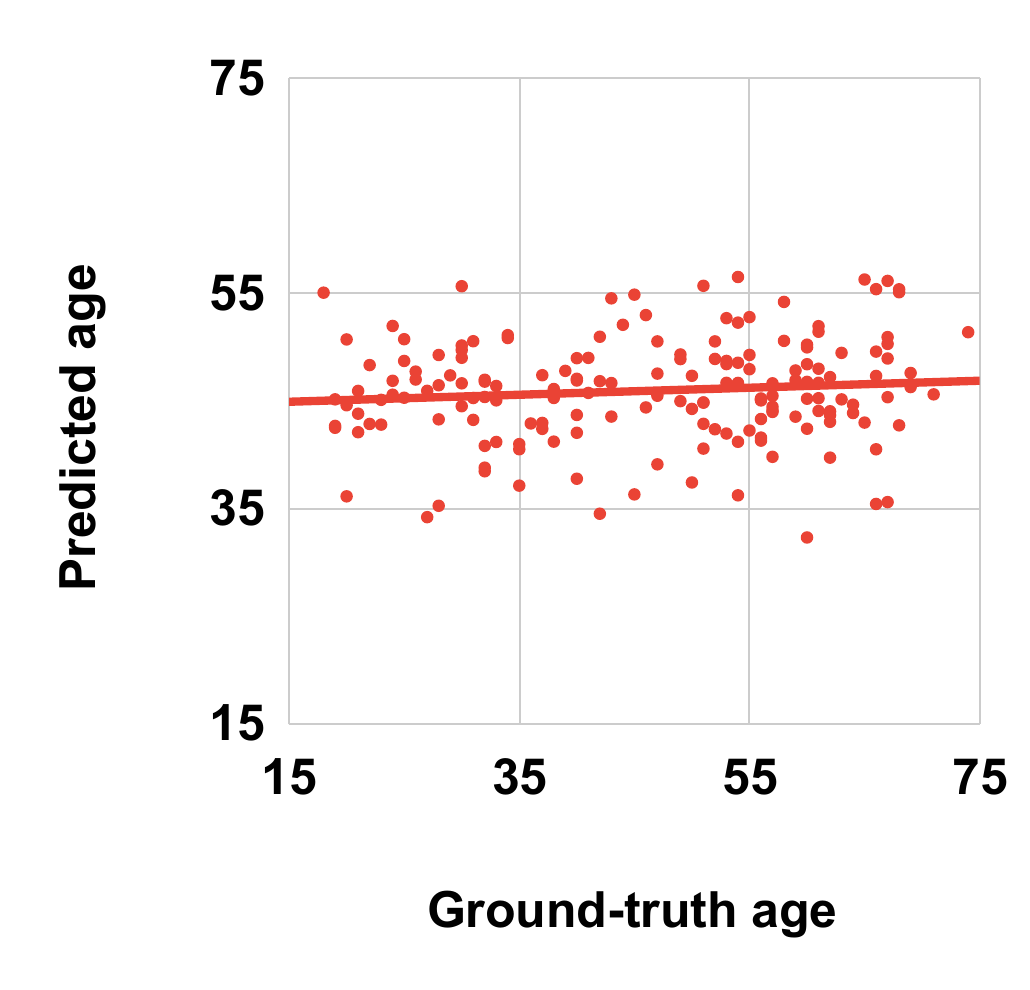}
\vspace{-5pt}
\caption{Age estimations with Marin+VM+ABM. The correlation between the ground-truth and predicted age is not significant ($r=0.11$, $p=0.16$ according to the Pearson coefficient). The results indicate successful mitigation of age bias.}
\vspace{-5pt}
\label{fig:age_estimation}
\end{figure}

In Figure \ref{fig:age_estimation}, we display age estimations with Marlin+VM+ABM for all subjects. These estimations appear random. Moreover, the Pearson correlation coefficient between the ground-truth and predicted age is $0.11$ with an associated P-value of $0.16$ which suggests the correlation is not significant. Both qualitative and quantitative results indicate successful mitigation of age bias.

\begin{figure}[t]
\centering
\footnotesize
    \begin{subfigure}{0.22\textwidth}
    \includegraphics[width=\textwidth]{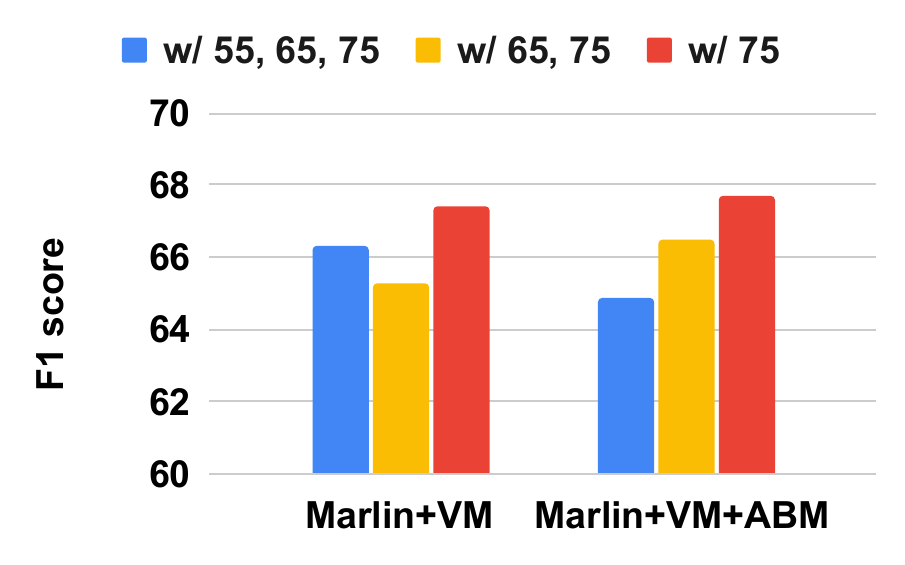}
    \vspace{-10pt}
    \caption{Performance for 5-fold cross-validation.}
    \vspace{-5pt}
    \label{fig:ablation_pretrain_cv}
    \end{subfigure}
    \begin{subfigure}{0.22\textwidth}
    \includegraphics[width=\textwidth]{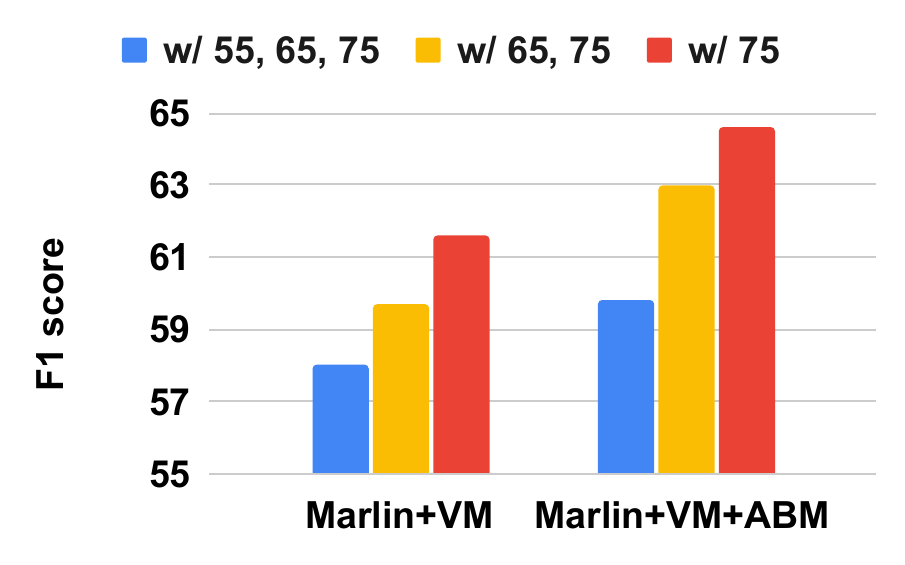}
    \vspace{-10pt}
    \caption{Average performance over different age groups.}
    \vspace{-5pt}
    \label{fig:ablation_pretrain_age}
    \end{subfigure}
    \caption{Performance of variation modeling with noisy segments sampled from different noise levels (55, 65, 75 dBA). Using noisy segments under 75 dBA noise level achieves the best performance.}
    \vspace{-10pt}
    \label{fig:ablation_pretrain}
\end{figure}

\noindent \textbf{Pre-train data for variation modeling.}
In VM, we have quiet segments with no background noise and noisy segments where noise levels are 55, 65, and 75 dBA. We evaluate two models (Marlin+VM and Marlin+VM+ABM) with three settings of pre-training, where noisy segments are sampled from (i) 55, 65, 75 dBA; (ii) 65, 75 dBA; (iii) 75 dBA. In Fig. \ref{fig:ablation_pretrain}, we find that 75 dBA achieves the highest performance for both models and evaluations. This finding is likely because subjects demonstrate the greatest variability in facial expressions between the quiet and 75dBA noise conditions, as compared to other noise conditions, providing more informative triplets for the pre-training phase.

\section{Conclusions}
In this paper, we explore a new problem: hearing loss detection from facial expressions in one-on-one conversations. We propose self-supervised variation modeling to capture the subject expression variability across noise levels. Our results show that ML models take advantage of age as a shortcut to the task, which hurts the performance of young subjects, so we mitigate the age bias via the use of a GRL. Extensive experiments and visualizations are conducted and the results demonstrate the effectiveness of our proposed method.

% \newpage
\bibliographystyle{IEEEbib}
\bibliography{ref}

\end{document}